\documentclass{article}

\usepackage{microtype}
\usepackage{graphicx}
\usepackage{subcaption}
\usepackage{booktabs} 

\usepackage{amsmath,amsfonts,bm}

\def\eqref#1{equation~\ref{#1}}

\def\1{\bm{1}}

\def\vm{{\bm{m}}}
\def\vn{{\bm{n}}}

\def\mA{{\bm{A}}}

\def\mD{{\bm{D}}}

\def\mI{{\bm{I}}}

\def\mL{{\bm{L}}}

\def\mS{{\bm{S}}}
\def\mT{{\bm{T}}}
\def\mU{{\bm{U}}}

\def\mW{{\bm{W}}}
\def\mX{{\bm{X}}}
\def\mY{{\bm{Y}}}
\def\mZ{{\bm{Z}}}

\DeclareMathAlphabet{\mathsfit}{\encodingdefault}{\sfdefault}{m}{sl}
\SetMathAlphabet{\mathsfit}{bold}{\encodingdefault}{\sfdefault}{bx}{n}

\def\gG{{\mathcal{G}}}

\def\sV{{\mathbb{V}}}

\usepackage[preprint]{icml2026}

\usepackage{amsmath}
\usepackage{amssymb}
\usepackage{mathtools}
\usepackage{amsthm}
\usepackage{float}
\usepackage{chngcntr}
\usepackage{tikz}

\usepackage[table]{xcolor}
\definecolor{StrongRed}{RGB}{255, 100, 100}      
\definecolor{MediumRed}{RGB}{255, 180, 180}      
\definecolor{LightRed}{RGB}{255, 220, 220}       

\usepackage{placeins}   
\usepackage{booktabs}   
\usepackage{multirow}
\usepackage{makecell}
\usepackage{amsmath}
\usepackage{amssymb}
\usepackage{mathtools}

\usepackage{siunitx}
\theoremstyle{plain}

\theoremstyle{definition}

\theoremstyle{remark}

\usepackage[textsize=tiny]{todonotes}\usepackage{xurl}      

\usepackage[table]{xcolor}
\usepackage{booktabs}
\usepackage{multirow}
\usepackage{colortbl}
\definecolor{headerblack}{RGB}{0,0,0}

\usepackage{xcolor}

\definecolor{level1}{RGB}{252,239,239}
\definecolor{level2}{RGB}{254,223,223}
\definecolor{level3}{RGB}{255,203,203}
\definecolor{level4}{RGB}{255,170,170}
\definecolor{level5}{RGB}{255,130,130}

\definecolor{red}{RGB}{255,0,0}      
\definecolor{purple}{RGB}{255,0,255} 
\definecolor{brightgreen}{RGB}{0,255,0} 
\definecolor{lightgreen}{RGB}{144,238,144} 

\usepackage{xurl}      
\usepackage{hyperref}  
\usepackage[capitalize,noabbrev]{cleveref}

\icmltitlerunning{Position: Spectral GNNs Are Neither Spectral Nor Superior for Node Classification}

\begin{document}

\twocolumn[
  \icmltitle{Position: Spectral GNNs Are Neither Spectral Nor Superior\\ for Node Classification }
 
  \icmlsetsymbol{equal}{*}

\begin{icmlauthorlist}
\icmlauthor{Qin Jiang}{1}
\icmlauthor{Chengjia Wang}{1}
\icmlauthor{Michael Lones}{1}
\icmlauthor{Dongdong Chen}{1}
\icmlauthor{Wei Pang}{1}

\end{icmlauthorlist}

  \icmlaffiliation{1}{Department of Computer Science, University of Heriot-Watt, Edinburgh, UK}

\icmlcorrespondingauthor{Qin Jiang}{qinjiang19871018@gmail.com}

  \icmlkeywords{GNN, Message Passing}

  \vskip 0.3in
]



\printAffiliationsAndNotice{}  

\begin{abstract}
Spectral Graph Neural Networks (Spectral GNNs) for node classification promise frequency-domain filtering on graphs, yet rest on flawed foundations. Recent work shows that graph Laplacian eigenvectors do not in general have the key properties of a true Fourier basis, but leaves the empirical success of Spectral GNNs unexplained. We identify two theoretical glitches: (1) commonly used ``graph Fourier bases'' are not classical Fourier bases for graph signals; (2) $(n-1)$-degree polynomials ($n$ = number of nodes) can exactly interpolate any spectral response via a Vandermonde system, so the usual ``polynomial approximation'' narrative is not theoretically justified. 
The effectiveness of GCN is commonly attributed to spectral low-pass filtering, yet we prove that low- and high-pass behaviors arise solely from message-passing dynamics rather than Graph Fourier Transform–based spectral formulations.
We then analyze two representative directed spectral models, MagNet and HoloNet. Their reported effectiveness is not spectral: it arises from implementation issues that reduce them to powerful MPNNs. When implemented consistently with the claimed spectral algorithms, performance becomes weak. This position paper argues that: \textbf{for node classification, Spectral GNNs neither meaningfully capture the graph spectrum nor reliably improve performance;} competitive results are better explained by their equivalence to MPNNs, sometimes aided by implementations inconsistent with their intended design.
\end{abstract}

\section{Introduction}
\label{sc:intro}

Graph Neural Networks (GNNs) for node classification encompass several active research branches, such as Message Passing Neural Networks (MPNNs)~\citep{gilmerNeuralMessagePassing2017, rossiEdgeDirectionalityImproves2023} and Spectral Graph Neural Networks (Spectral GNNs)~\citep{zhangMagNetNeuralNetwork, BernNet}.
Spectral GNNs form a niche branch rooted in spectral graph theory \citep{chung1997spectral} and polynomial approximation techniques \citep{hammondWaveletsGraphsSpectral2009a}.
However, their polynomial filters are problematic in that performance does not reliably improve as filter order $K$ increases~\citep{BernNet}. 
A recent position paper~\citep{guoPositionSpectralGNNs2025} further argues that graph Laplacian eigenvectors should not be treated as a canonical Fourier basis in the classical sense, thereby challenging the usual “spectral” interpretation. We agree with this conclusion. 
However, this limitation should not be attributed to a continuous-versus-discrete mismatch claimed by \citet{guoPositionSpectralGNNs2025}: Fourier representations extend naturally to discrete settings via the DFT~\citep{oppenheim99, gonzalezDigitalImageProcessing2017}. 
Instead, the gap stems from structural differences between Euclidean domains and general graphs—most notably the absence of a global shift/translation operator—so the Laplacian eigenbasis lacks some of the canonical properties that underpin Fourier analysis in regular domains.

\textbf{We take a stronger stance: Spectral GNNs are theoretically flawed with several pitfalls, and their seemingly strong empirical performance stems primarily from equivalences to simpler MPNNs, rather than from any genuinely spectral mechanism.}

\subsection{Structure of this Position Paper}

\textbf{Section~\ref{sec:graph_fourier}:} We prove that
Graph Fourier Basis $\neq$ Fourier Basis of Graph, exposing our first theoretical concern.

\textbf{Section~\ref{sec:poly_appro}:}
Even granting that graph Fourier bases qualify as true Fourier bases, polynomial approximations fail on their own terms.
Following Spectral GNN reasoning, 
we show that on an $n$-node graph, an $(n\!-\!1)$th-order polynomial exactly interpolates any ideal spectral filter via a Vandermonde system, yielding a determined interpolation rather than a genuine approximation.
Moreover, we point out that a $k$th-order polynomial exactly recovers $k$-hop message passing, explaining the effectiveness of Spectral GNNs as an equivalence to simpler MPNNs.

\textbf{Section~\ref{sc:low_pass}:} We show that the commonly cited low-/high-pass behavior of GNN spectral filters can be justified via message passing; in contrast, a Graph Fourier Transform-based argument by itself cannot establish this filtering claim.

\textbf{Section~\ref{sec:mag_holo}:} We analyze two prominent Spectral GNNs for directed graphs---MagNet~\citep{zhangMagNetNeuralNetwork} and HoloNet~\citep{kokeholonets}. Their strong empirical results stem not from spectral mechanisms, but from implementation bugs that reduce them to much simpler MPNNs.

\paragraph{Scope and Terminology} 
We focus on node classification, primarily on general graphs (with directed and undirected graphs as the special case where each edge is accompanied by its reverse).
Let \( \gG=(\sV,\mathcal{E}) \) be a graph with \(n=\lvert\sV\rvert\) nodes.
Node features are stored in \( \mX \in \mathbb{R}^{n \times f} \), and node labels are \( y_i \in \{1,\ldots,c\} \).
The adjacency matrix is \( \mA \in \{0,1\}^{n \times n} \) with \( \mA_{ij}=1 \) indicating an edge \(i \to j\).
Let \( \mD \) denote a degree matrix (specified as in-/out-degree when needed).
For undirected graphs, the Laplacian is \( \mL = \mD - \mA \), and the normalized Laplacian is \( \hat{\mL} = \mI - \mD^{-1/2}\mA\mD^{-1/2} \).
We use \( \hat{\mA} \) to denote a normalized adjacency (defined in context).


\section{Graph Fourier Basis}
\label{sec:graph_fourier}

\subsection{Background}
\label{sec:bk_fourier}

Spectral graph learning builds upon (i) Graph Fourier Basis as the conceptual foundation and (ii) polynomial approximations (e.g., Chebyshev, Bernstein) as a practical surrogate to avoid eigen-decomposition \citep{guoPositionSpectralGNNs2025}. 
A Graph Fourier Transform (GFT) is typically defined through the eigen-decomposition of the normalized graph Laplacian $\hat\mL=\mI-\hat{\mA}$:
\begin{equation}
\label{eq:L_decom}
\hat\mL = \mU \bm{\Lambda} \mU^\top ,
\end{equation}
where $\mU$ contains orthonormal eigenvectors and $\bm{\Lambda}$ is diagonal. $\hat{\mA}$ is normalised adjacency matrix.

\subsection{Graph Fourier Basis $\neq$ Fourier Basis of Graph}
While $\mU$ indeed forms an orthonormal basis and defines a valid transform for the \emph{operator} $\mL$, identifying it as the graph’s own Fourier basis is conceptually misguided. 

The Fourier transform was originally developed for one-dimensional signals (e.g., vibrating strings and time sequences) \citep{steinFourierAnalysisIntroduction2003}.
Within Euclidean domains, the notion extends naturally to higher dimensions. For example, for images on $\mathbb{Z}^2$, the 2D Fourier transform represents an image as a superposition of sinusoidal patterns that vary along the horizontal and vertical directions of the grid \citep{russ2016imageprocessinghandbook, gonzalezDigitalImageProcessing2017}.
More fundamentally, Fourier modes in Euclidean settings arise as eigenfunctions (characters) of the translation group; 
consequently, “frequency” is tied to translation invariance and to sinusoidal patterns that are globally meaningful across the domain.
On discrete domains the same principle holds: the discrete Fourier transform diagonalizes the action of cyclic shifts on $\mathbb{Z}_n$, and its modes are characters that represent globally meaningful oscillations on a regular grid \citep{oppenheim99}.

For general graphs, however, there is no canonical notion of translation and no globally consistent set of orthogonal directions; thus, the classical semantics of frequency do not directly carry over \citep{ricaudFourierCouldBe2019}. 
Recent discussions further stress that the semantic properties that make Fourier analysis powerful in Euclidean settings may not be preserved when this terminology is transferred to arbitrary graphs \citep{guoPositionSpectralGNNs2025}.

Using Laplacian eigenvectors as the graph’s “Fourier basis” is analogous to the following construction in the image domain: 
rather than applying the standard 2D Fourier transform—whose modes are global sinusoids aligned with the horizontal and vertical directions—one instead diagonalizes a pixel-connectivity Laplacian and treats its eigenvectors as “spectral components.” This procedure indeed produces an orthonormal basis and defines a valid transform for that Laplacian. However, it is conceptually distinct from the classical Fourier basis, whose modes correspond to oscillations along the grid’s translation directions. Consequently, the resulting coefficients do not, in general, admit the classical interpretation of a “spectrum” tied to translation-invariant oscillations. This is a pitfall: Graph Fourier Basis $\neq$ Fourier basis of graph.

Recent work~\citep{guoPositionSpectralGNNs2025} reaches a similar conclusion but misattributes the issue to continuous-vs-discrete mismatch, overlooking the well-established Discrete Fourier Transform~\citep{oppenheim99}. The real issue is that graphs generally lack the translation structure that underlies classical Fourier analysis.


\section{Polynomial Approximations}
\label{sec:poly_appro}

\subsection{Background}
\label{sec:bk_poly}

While interpreting Laplacian eigenvectors as the graph’s “Fourier basis” remains conceptually debatable, the subsequent use of polynomial approximations raises additional issues even granting the Graph Fourier Basis interpretation. 

\begin{figure}[ht]
  \vskip 0.2in
  \begin{center}
    \centerline{\includegraphics[width=\columnwidth]{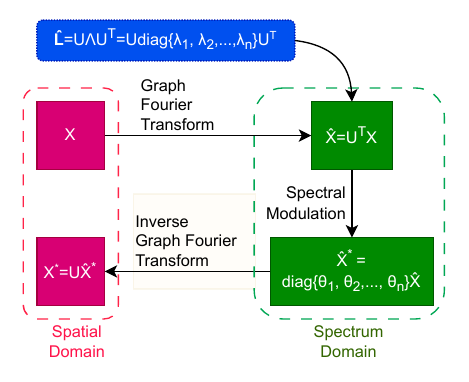}}
\caption{Graph Fourier transform and its inverse, showing spectral modulation by a diagonal filter in the Fourier domain.}

    \label{fig:spectrum_GNN}
  \end{center}
\end{figure}

Spectral GNNs take node features $\mX \in \mathbb{R}^{n \times d}$ as input signals on $d$ channels, map them to the spectral domain via the Graph Fourier Transform, apply a spectral modulation $g(\lambda)$, and then return to the spatial domain using the inverse transform \citep{guoGraphNeuralNetworks2023}, as shown in Figure \ref{fig:spectrum_GNN}. 

To avoid the eigendecomposition of Equation \ref{eq:L_decom}, the filter 
$g(\lambda)$ is typically approximated by a truncated polynomial $h(\lambda)=\sum_{k=0}^{K} \alpha_k \lambda^k$, so that the filtered signal can be implemented using powers of the normalized Laplacian \citep{kipfSemiSupervisedClassificationGraph2017,wangHowPowerfulAre2022} as follows:
\begin{equation}
\label{eq:X_poly}
\mX^* \approx \sum_{k=0}^{K} a_k \hat\mL^k \mX.
\end{equation}

\subsection{Pitfalls of Truncated Polynomial Approximations}

Consider an “ideal” spectral modulation $\bm{\theta}^*=(\theta_1^*, ..., \theta_n^*)$ on a graph with eigenvalues $\lambda_1, ..., \lambda_n$. Writing the polynomial constraints
\begin{equation}
\label{eq:n_linear}
    {\theta_i}^* = \sum_{k=0}^{K}  {\alpha_k} {\lambda_i}^k, i=1,...,n,
\end{equation}
yields a linear system in the coefficients $\alpha_0$, ..., $\alpha_K$. 
When the eigenvalues $\lambda_i$ are distinct, setting $K=n-1$ produces a square Vandermonde system with a unique solution of $\bm\alpha=(\alpha_0, ..., \alpha_K)$: there is exactly one polynomial of degree at most $n-1$ interpolating the $n$ target values ${\bm\theta}^*$ exactly. 
Increasing to $K\geq n$ only introduces unnecessary underdetermination without improving the fit on this fixed graph. 
An "infinite-hop" polynomial is therefore never necessary: on any fixed graph, $n-1$ hops suffice to realise an arbitrary spectral filter exactly.

Crucially, this constitutes a finite interpolation problem rather than a traditional polynomial approximation task. Classical approximation techniques—such as Taylor or Chebyshev expansions—progressively improve as terms of increasing degree refine the estimate of a continuous function $f$. 
In contrast, interpolation over a Vandermonde system lacks this hierarchical structure: all polynomial orders are equally important. 
Once the coefficients $\bm\alpha$ are determined, each coefficient $\alpha_k$-regardless of degree, whether multiplying $\lambda^0$ or $\lambda^{n-1}$-contributes equally to matching $\bm{\theta}^* $ on the spectrum. 

However, empirical results in previous studies, such as \citet{jiang2025demystifyingmpnnsmessagepassing}, are inconsistent with this theoretical deduction. Experiments consistently show that aggregating from low-order neighborhoods outperforms aggregation from higher-order ones — the opposite of what the spectral motivation would suggest. This inconsistency calls into question the validity of the theoretical framework from which the deduction was derived, and provides further evidence that the Graph Fourier Transform is a flawed foundation for modeling graph signals.
This theoretical deduction also sits in tension with a recurring empirical trend: truncated polynomial approximations inadvertently favour low-degree coefficients $\alpha_0$, $\alpha_1$, ... (corresponding to local neighborhood aggregation), while under-utilizing higher-degree terms.

We now follow further deductions stemming from this flawed premise. 
Spectral GNNs that propose alternative polynomial families, such as Chebyshev expansions \citep{defferrardConvolutionalNeuralNetworks2016}, first-order Chebyshev expansion \citep{kipfSemiSupervisedClassificationGraph2017}, Bernstein bases \citep{BernNet}, or others \citep{newtonPolynomial}, amounts to blindly exploring different values of $\alpha_k$.
For example, ChebNet \citep{defferrardConvolutionalNeuralNetworks2016}, effectively sets $\alpha_0$=1, $\alpha_1$=1, while GCN \citep{kipfSemiSupervisedClassificationGraph2017} corresponds to $\alpha_0$=2, $\alpha_1$=-1 (after renormalization).
These are merely particular points in the overall interpolation space, without any structurally privileged status.
Furthermore, Equation \ref{eq:n_linear} is inherently dataset-specific: $\bm\Lambda$ and $\bm\theta$ vary across graphs. Spectral GNNs thus imply dataset-specific models, undermining their claimed generality. 
Collectively, these deductions highlight an internal inconsistency: conclusions derived from the theoretical formulation of spectral GNNs contradict their empirical practice.

In summary, we expose the limitations of spectral GNNs from multiple perspectives: (1) each of the two constituent steps is independently problematic, rooted in a misapplication of Fourier analysis and polynomial approximation respectively.
(2) theoretical implications contradict empirical evidence that low-order terms are more important than higher-order terms; and (3) the design choices of spectral GNNs, such as favouring low-order terms, unprincipled parameter search, and implicit dataset-specificity that undermines their claimed generality, are inconsistent with their own theoretical deductions.

\subsection{What Polynomial Approximation Truly Leads To}

The two conceptual flaws identified above — the graph Fourier basis interpretation \citep{guoPositionSpectralGNNs2025} and the polynomial approximation pitfalls — turn out, in combination, to be accidentally productive. 
With their combination, spectral filters are polynomial functions of aggregation via graph Laplacian; and since the Laplacian encodes local adjacency, different polynomial filters correspond to different weighted combinations of spatial aggregations, with polynomial order directly controlling the maximum hop distance \citep{balcilar2021analyzing}. 
Thus, although each ingredient is internally flawed, together they converge on a correct and interpretable computational primitive: the MPNN.

Specifically, polynomial filters in Equation~\ref{eq:X_poly}—where $\hat\mL^k \mX$ aggregates $k$-hop neighborhood information \citep{jiang2025demystifyingmpnnsmessagepassing}—explicitly correspond to spatially localized, weighted combinations of neighborhood aggregation that avoid eigendecomposition altogether \citep{balcilar2021analyzing}.
ChebNet \citep{defferrardConvolutionalNeuralNetworks2016} and GCN \citep{kipfSemiSupervisedClassificationGraph2017} are canonical examples: both yield bounded-hop spatial computations despite their spectral motivation.

\medskip
The limitations of Spectral GNNs extend beyond the pitfalls discussed above; Appendix~\ref{sec:fur_dis} provides additional reasons why they are ill-suited to node classification.



\section{Low-Pass and High-Pass Filters}
\label{sc:low_pass}

In the discrete-time Fourier setting, consider the linear time-invariant (LTI) system
\[
y[n] = x[n] + x[n-1].
\]
For a complex exponential input $x[n]=e^{i\omega n}$, the output has the form
$y[n]=H(\omega)e^{i\omega n}$, where the frequency response is
\[
H(\omega)=1+e^{-i\omega}.
\]
Its magnitude is largest at low $\omega$ and smallest at high $\omega$, hence the system is \emph{low-pass} \citep{oppenheim99}. A detailed derivation is provided in Appendix~\ref{sec:dft_low_pass}.
Similarly, the system
\[
y[n] = x[n] - x[n-1]
\]
has frequency response $H(\omega)=1-e^{-i\omega}$, whose magnitude is smallest at low $\omega$ and largest at high $\omega$, and is therefore \emph{high-pass}. Intuitively, $x[n]+x[n-1]$ emphasizes slowly varying (smooth) components, while $x[n]-x[n-1]$ emphasizes rapid changes.

\subsection{GCN as a Low-Pass Filter}
Analogous notions of low-pass and high-pass filtering arise as spectral filters \citep{balcilar2021analyzing, liu2025graph}. 
It is widely recognized that graph convolutional networks (GCNs) exhibit a low-pass bias \citep{wuSimplifyingGraphConvolutional2019, chen2024polygcl}, often attributed to aggregation with self-loops (i.e., using $\tilde{\mA}=\mA+\mI$). However, a commonly cited justification based solely on graph Laplacian eigenvalue ordering does not by itself establish a \emph{filtering} claim.

\citet{wuSimplifyingGraphConvolutional2019} argue as follows. After adding self-loops, the eigenvalues of
\[
\tilde{\mL}=\mI-\tilde{\mD}^{-1/2}\tilde{\mA}\tilde{\mD}^{-1/2}
\quad\text{and}\quad
\mL=\mI-\mD^{-1/2}\mA\mD^{-1/2}
\]
(with $\tilde{\mD}$ and $\mD$ the degree matrices of $\tilde{\mA}=\mA+\mI$ and $\mA$, respectively) satisfy an ordering relationship; in particular, they prove that the largest eigenvalue of the normalized Laplacian decreases after adding self-loops. Based on this eigenvalue shrinkage, they interpret GCNs with self-loops as exhibiting a low-pass-type behavior.

However, this reasoning does not by itself establish a \emph{filtering} claim. The above result is a statement about the \emph{spectrum of the Laplacian}. By contrast, to call an operator ``low-pass'' one must characterize its \emph{frequency response}, i.e., how the \emph{propagation operator} scales Fourier modes across Laplacian eigenvalues (preferentially amplifying low-frequency modes and attenuating high-frequency modes). 
An ordering or shrinkage of Laplacian eigenvalues 
influences the scaling behavior of matrix‑Laplacian multiplication, 
but does not imply that the corresponding GCN propagation operator has maximal gain on low-frequency components.

We agree with the qualitative conclusion that GCNs are low-pass biased, but we contend that it does not follow directly from Laplacian eigenvalue ordering. Intuitively, using $\mA+\mI$ promotes feature similarity among neighboring nodes and thus favors smooth node signals. In this sense, the pair $(\mI+\mA)$ and $(\mI-\mA)$ is analogous to the classical low-pass and high-pass constructions in the discrete-time Fourier setting.

\subsection{A Hop-Domain View: Shift-Invariant Filtering on Graphs}

We now introduce a hop-indexed viewpoint that parallels the classical theory of LTI systems in discrete-time signal processing. The key difference is that the index $n$ below is \emph{not physical time}, but the \emph{hop depth} of message passing. Concretely, let $\mS$ denote a fixed graph propagation (shift) operator (e.g., $\mS=\tilde{\mD}^{-1/2}\tilde{\mA}\tilde{\mD}^{-1/2}$ for GCN, or another choice depending on the model). Given node features $\mX$, we define a hop-indexed sequence \citep{jiang2025demystifyingmpnnsmessagepassing}
\[
\mX[0] = \mX,
\qquad
\mX[n] = \mS^{n}\mX \ \ (n\ge 1),
\]
so that $\mX[n]$ aggregates information from $n$ successive propagations.

Under this interpretation, we can study \emph{linear shift-invariant} operators acting on the discrete index $n$ (the hop index), in direct analogy with discrete-time LTI systems. In particular, consider the hop-domain operators
\begin{equation}
\label{eq:low_pass_graph}
    \mY[n] = \mX[n] + \mX[n-1],
\end{equation}
and
\begin{equation}
\label{eq:high_pass_graph}
    \mY[n] = \mX[n] - \mX[n-1].
\end{equation}
These operators are linear and \emph{shift-invariant with respect to $n$}: shifting the input sequence $\{\mX[n]\}$ along the hop index results in the same shift of the output sequence $\{\mY[n]\}$. Hence, they admit the standard frequency-response interpretation from discrete-time LTI theory, with the hop index playing the role of the discrete-time variable.

Moreover, this hop-domain construction yields an immediate interpretation of common GNN updates. For example, the GCN propagation $(\mI+\mS)\mX$ can be written as
\[
(\mI+\mS)\mX = \mX[0] + \mX[1],
\]
which is exactly the $n=1$ instance of the hop-domain smoothing operator in Equation~\ref{eq:low_pass_graph}. Thus, a single message-passing step with a residual/self-loop term implements a first-order hop-domain smoothing operation, providing an alternative and transparent route to the widely observed low-pass bias of GCN-type layers.

Importantly, this argument is conceptually different from reasoning based on the Graph Laplacian Transform; the hop-domain view characterizes low-/high-pass behavior directly through shift-invariant operations on propagation depths.

In sum, the highly cited claim that ``spectral filters'' are low-pass or high-pass cannot be deduced from Graph Laplacian Transform-based spectral GNN theories; rather, it follows from message passing, viewed as hop-invariant filtering analogous to time-invariant filtering.

\section{Case Study: Popular spectral GNNs for Directed Graphs}
\label{sec:mag_holo}

\paragraph{MPNN is the reason for GCN's success} 
Among spectral GNN models for undirected graphs, GCN \citep{kipfSemiSupervisedClassificationGraph2017} has become one of the classical and powerful baselines \citep{luoClassicGNNsAre2024}. Although it claims to be based on a first-order Chebyshev polynomial approximation of spectral convolutions, the actual formulation jumps directly to setting $\alpha_0 = 2$ and $\alpha_1 = -1$ (after renormalization). 
Since a first-order Chebyshev polynomial would give  $\alpha_0 = 1$ and $\alpha_1 = 1$, strictly speaking, GCN is not first-order Chebyshev polynomial approximation of spectral convolutions, and thus cannot be viewed as a purely spectral GNN. This coefficient choice is not supported by the spectral-GNN theory itself; instead, its effectiveness is best explained through a spatial-domain perspective: 1-hop neighborhood aggregation with self-loops \citep{balcilar2021analyzing}.

\paragraph{Directed spectral GNNs: MagNet and HoloNet} 
We now examine two of the most popular spectral GNNs for directed graphs—MagNet \citep{zhangMagNetNeuralNetwork} and HoloNet \citep{kokeholonets}. Despite their theoretical designs rooted in spectral operators for directed Laplacians, we reveal that their strong empirical performance stems from implementation flaws. These coding deviations contradict the intended spectral formulations and, in practice, transform the models into conventional spatial aggregators. This mirrors the phenomenon observed in GCN, where the “spectral” label masks a fundamentally message-passing architecture.

\subsection{MagNet = GraphSAGE with GCN normalisation}
\label{mag=sage}

\subsubsection{Spectral GNN for Directed Graphs}
\label{sc:bk_spec_direct}

The adjacency matrix of a directed graph \citep{tongDigraphInceptionConvolutional2020} is asymmetric, which prevents direct spectral eigendecomposition of its Laplacian. 
While this should reveal the incompatability of Spectral Graph Theory (which deals with identical nodes and bidirected edges) with modern graph learning, MagNet \citep{zhangMagNetNeuralNetwork} push it forward by introducing a complex-valued {Hermitian} Laplacian: directionality is encoded via complex phases while the matrix remains Hermitian.

The adjacency matrix of a directed graph is generally asymmetric \citep{tongDigraphInceptionConvolutional2020}, so the associated Laplacian is non-symmetric and does not admit the standard real orthogonal spectral eigendecomposition that underpins classical spectral graph theory. 
This is not a minor technicality: it reflects a structural mismatch between spectral graph theory—developed for undirected graphs with bidirected edges and identical nodes—and modern graph learning, where edges are often directed and node features are diverse.
Rather than treating this mismatch as a signal to move beyond the spectral paradigm, MagNet \citep{zhangMagNetNeuralNetwork} attempts to preserve it by introducing a complex-valued Hermitian Laplacian, encoding edge direction through complex phases while maintaining Hermitian symmetry to recover a spectral decomposition.

As a result, its eigenvalues are real and eigenvectors form an orthonormal basis, enabling a spectral-style framework for directed graphs.
In this section, we show that MagNet is mathematically equivalent to GraphSAGE \citep{hamiltonInductiveRepresentationLearning2018} equipped with GCN-style normalisation.

\subsubsection{Hermitian Construction in MagNet}
For a directed graph, the adjacency matrix $\mA$ is generally not equal to its transpose $\mA^\top$. 
MagNet \citep{zhangMagNetNeuralNetwork} first symmetrises the adjacency via
\[
\mA_s = \frac{1}{2}(\mA + \mA^\top),
\]
then applies GCN normalisation to obtain $\widetilde{{\mA}_s}$.
Directionality is reintroduced through an element-wise complex phase matrix
\[
\bm\Theta = e^{2 \pi q j (\mA-\mA^\top)},
\]
where $q\in\mathbb{R}$ controls the phase magnitude and $j^2=-1$.
The resulting complex adjacency used by MagNet is
\[
\resizebox{\linewidth}{!}{$
\begin{aligned}
\widehat{{\mA}_s}
&= \widetilde{{\mA}_s} \odot \bm{\Theta}
 = \widetilde{{\mA}_s} \odot e^{2\pi q j(\mA-\mA^\top)} \\
&= \widetilde{{\mA}_s} \odot \cos\!\big(2\pi q(\mA-\mA^\top)\big)
 + j\,\widetilde{{\mA}_s} \odot \sin\!\big(2\pi q(\mA-\mA^\top)\big),
\end{aligned}
$}
\]
where $\odot$ denotes element-wise multiplication.
Equivalently, $\widehat{{\mA}_s}$ admits the decomposition
\[
\widehat{{\mA}_s} = \Re(\widehat{{\mA}_s}) + j\,\Im(\widehat{{\mA}_s}),
\]
with
\[
\Re(\widehat{{\mA}_s})
= \widetilde{{\mA}_s} \odot \cos\!\big(2\pi q(\mA-\mA^\top)\big),
\]
\[
\Im(\widehat{{\mA}_s})
= \widetilde{{\mA}_s} \odot \sin\!\big(2\pi q(\mA-\mA^\top)\big).
\]

Each entry of $\Re(\widehat{{\mA}_s})$ and $\Im(\widehat{{\mA}_s})$ takes a value in $\{1,0,\cos\alpha,\sin\alpha\}$, where $\alpha = 2\pi q$, as shown in Table~\ref{tab:MagNet_Hermitian}.

As a result, $\Re(\widehat{{\mA}_s})$ — denoted $\overline{{\mA}_s}$ — is a real symmetric matrix matching $\widetilde{{\mA}_s}$ on bidirectional edges, while for unidirectional edges the corresponding entries are scaled by a factor of $\cos\alpha$ relative to those of $\widetilde{{\mA}_s}$. $\Im(\widehat{{\mA}_s})$ is skew-symmetric, and can be written as
\begin{equation}
\label{As_imag}
\Im(\widehat{{\mA}_s})
= \tfrac{1}{2}\sin\alpha \mD^{-\frac{1}{2}}(\mA-\mA^\top)\mD^{-\frac{1}{2}},
\end{equation}
where $\mD$ is the degree matrix of the graph.

\begin{table}[ht]
\centering
\captionsetup{font=normal}
\caption{Case enumeration of the elements with entry $mn$ in adjacency matrices for different edge types between node $\vm$ and node $\vn$. Here, $\mA_s$ is the symmetrized adjacency matrix, $\widetilde{{\mA}_s}$ is its GCN-normalized adjacency matrix, and $\widehat{{\mA}_s}$ is the complex-valued adjacency matrix used in MagNet with parameter $\alpha=2\pi q$. The variable $d$ denotes the node degree.}
\label{tab:MagNet_Hermitian}

\resizebox{\linewidth}{!}{%
\begin{tabular}{c|c|c|c|c|c}
\toprule
\multicolumn{6}{c}{\textbf{MagNet} ($\alpha=2\pi q$)} \\
\midrule
\textbf{Edges} & ${\mA}_s$ & $\widetilde{{\mA}_s}$ & $\widehat{{\mA}_s}$ & $\Re(\widehat{{\mA}_s})$ & $\Im(\widehat{{\mA}_s})$ \\
\midrule
\(\vm \rightarrow \vn\)
& $0.5$
& $\frac{0.5}{d}$
& $\frac{0.5}{d} e^{j\alpha}$
& $\frac{0.5}{d}\cos\alpha$
& $\frac{0.5}{d}\sin\alpha$ \\
\midrule
\(\vn \rightarrow \vm\)
& $0.5$
& $\frac{0.5}{d}$
& $\frac{0.5}{d} e^{-j\alpha}$
& $\frac{0.5}{d}\cos\alpha$
& $-\frac{0.5}{d}\sin\alpha$ \\
\midrule
\(\vm \leftrightarrow \vn\)
& $1$
& $d^{-1}$
& $d^{-1}$
& $d^{-1}$
& $0$ \\
\midrule
\(\vm \not\leftrightarrow \vn\)
& $0$ & $0$ & $0$ & $0$ & $0$ \\
\bottomrule
\end{tabular}%
}
\end{table}

\subsubsection{Hermitian Propagation and Output Derivation}

In MagNet, node features are initialised as $\widehat \mX = \mX+j\mX$.
However, when following the ChebNet recursion \citep{defferrardConvolutionalNeuralNetworks2016}, a coding error was introduced: an extra subtraction of $\mI$ occurred when constructing the Laplacian $\hat\mL=\mI -\widehat{ \mA_s}$. 
This caused a mistaken recurrence:
\begin{equation}
    \widetilde{\mT_{k+2}} = 2\widetilde \mL \mT_{k+1} - \mT_{k}, 
    \text{where} \widetilde {\mL} =\mI -\widehat{ \mA_s}-\mI=-\widehat{ \mA_s}
\end{equation}
rather than the correct Chebyshev form: 
\begin{equation}
    \widetilde{\mT_{k+2}} = 2\hat \mL \mT_{k+1} - \mT_{k}
\end{equation}

When $K = 1$, the Hermitian output is as follows:
\begin{equation}
\label{out1}
\begin{aligned}
\mZ_1 &= \sigma(\widehat {\mT}_1 \mX  \mW_{11})=  \sigma \big((\mI+ j\bm0)(\mX+j\mX)\mW_{11} \big) \\
&= \sigma \big((\mX+j\mX)\mW_{11} \big)
=  \sigma (\mX)+j \sigma(\mX\mW_{11} \big),
\end{aligned}
\end{equation}
where $\mW_{11}$ is a learnable weight matrix, and $\sigma(\cdot)$ is a nonlinear activation function; in MagNet, $\sigma(\cdot)=\mathrm{ReLU}(\cdot)$. 
For simplicity, we omit $\sigma(\cdot)$ in the following derivations, as this does not affect the final conclusion of our proof \citep{HORNIK1991251,HornikKurt1989Mfna,jiang2026scaleaware,xuHowPowerfulAre2019,wuSimplifyingGraphConvolutional2019}.

When $K=2$, substituting $\mZ_1$ from Equation~\ref{out1} into the $K=2$ update, the complex-valued output becomes:
\begin{equation}
\begin{split}
 \mZ_2 &= \sigma \big(\mZ_1 +   \big(   \Re(\widetilde{{\mT}_{2}})  + j\Im(\widetilde{{\mT}_2})\big)(\mX+j\mX)\mW_{21} \big) \\
&
= \sigma \big( \mZ_1 +  \big(-\Re(\widehat{{\mA}_s}) - j\Im(\widehat{{\mA}_s})\big)(\mX+j\mX)\mW_{21} \big) \\
&  = \sigma \big( \mX\mW_{11}+\Im(\widehat{{\mA}_s})  \mX\mW_{21} - \Re(\widehat{{\mA}_s}) \mX \mW_{21} \\
&
 + j\big( \mX\mW_{11}- \Re(\widehat{{\mA}_s})  \mX \mW_{21} - \Im(\widehat{{\mA}_s})  \mX \mW_{21}\big) \big),
\end{split}
\label{out2}
\end{equation}
where $\mW_{21}$ is also a learnable weight matrix.

In the end, MagNet concatenates the real and imaginary parts of the Hermitian output. 

For $K=1$, by Equation~\ref{out1} we have $\Re(\mZ_1)=\Im(\mZ_1)=\mX\mW_{11}$. Substituting into the readout yields
\begin{flalign}
\label{eq:out1_total}
Out_1
&= \sigma \big(\Re(\mZ_1)\mW_{01} + \Im(\mZ_1)\mW_{02} \big) \notag \\
&= \sigma \big( \mX\mW_{11}\mW_{01} + \mX\mW_{11}\mW_{02} \big) \notag\\
&=\sigma ( \mX\mW_{1} + \mX\mW_{2}),
\end{flalign}
where $\mW_{1}=\mW_{11}\mW_{01}$ and $\mW_{2}=\mW_{11}\mW_{02}$ are learnable weight matrices.

For $K=2$, based on Equation~\ref{out2}, we obtain:
\begin{equation} 
\label{eq:out2_q=o}
\begin{split}
   &Out_2  =  \sigma \big( \Re(\mZ_2)\mW_{22} + \Im(\mZ_2) \mW_{23}  \big)\\
 &= \sigma \Big( \big( \mX\mW_{11}+\Im(\widehat{{\mA}_s})  \mX\mW_{21} - \Re(\widehat{{\mA}_s}) \mX \mW_{21} \big)\mW_{22} + \\
&  \big( \mX\mW_{11}- \Re(\widehat{{\mA}_s})  \mX \mW_{21} - \Im(\widehat{{\mA}_s})  \mX \mW_{21} \big) \mW_{23}  \Big)\\
& =  \sigma \big(\mX\widehat \mW_1+\Re(\widehat{{\mA}_s}) \mX\widehat \mW_2+\Im(\widehat{{\mA}_s})\mX \widehat \mW_3 \big), 
\end{split}
\end{equation}
where 
\begin{flalign}
&\widehat{\mW}_1 := \mW_{11}(\mW_{22}+\mW_{23}),\notag\\
&\widehat{\mW}_2 := -\mW_{21}(\mW_{22}+\mW_{23}),\notag\\
&\widehat{\mW}_3 := \mW_{21}(\mW_{22}-\mW_{23}) \notag
\end{flalign}
are all learnable weight matrices.

Substituting Equation~\ref{As_imag}, Equation~\ref{eq:out2_q=o} can further be simplified to :
\begin{equation}
\begin{split}
&Out_2
= \sigma \big( \mX\widehat\mW_1
+ \mD^{-\frac{1}{2}}\overline{\mA_s}\mD^{-\frac{1}{2}}\mX\widehat\mW_2 \\
&\quad + \tfrac{1}{2}\sin \alpha\,
\mD^{-\frac{1}{2}}(\mA-\mA^\top)\mD^{-\frac{1}{2}}\mX\widehat\mW_3 \big).
\end{split}
\label{eq:out3}
\end{equation}

In MagNet’s Chebyshev-polynomial implementation, the polynomial order is set to $K=2$ \citep[Supplementary Material]{zhangMagNetNeuralNetwork}; therefore, Equation~\ref{eq:out3} gives the layer-wise output of MagNet.

\subsubsection{Relationship to GraphSAGE}

\begin{table}[htbp]

    \captionsetup{font=normal} 
\caption{Classification accuracy (\%) of Dir-GNN \cite{rossiEdgeDirectionalityImproves2023} and MLP with different feature configurations and normalization schemes on Telegram datasets. Feature configurations include: original node features from datasets (Origin Feature), constant features (No Feature, all set to 1), and node degree variants (in-degree, out-degree, or both). \textbf{Bold} value indicates learning failure with row normalization and no features. \underline{Underline} value indicates that predictions can be made accurately using only node degrees by MLP without message passing.}
    \label{tab:tel_row_norm}
    \centering
    \setlength{\tabcolsep}{2pt}
    \vskip 0.15in
\begin{center}
\begin{small}
\begin{tabular}{lc|cccc}
\toprule
\textbf{Telegram}  & \textbf{MLP}&\textbf{None} &  \textbf{Row} & \textbf{Sym} & \textbf{Dir}  \\
      \midrule

Origin&
\cellcolor{level1} 38.0±7.2&
\cellcolor{level5} 95.6±2.8&
\cellcolor{level2} 74.2±5.5&
\cellcolor{level4} 93.0±4.1&
\cellcolor{level4} 92.8±4.7\\

No Feature&
\cellcolor{level1} 38.0±0.0&
\cellcolor{level5} 95.4±4.0&
\cellcolor{level1} \textbf{38.0±0.0}&
\cellcolor{level4} 93.0±4.7&
\cellcolor{level4} 93.0±3.0\\

In-degree&
\cellcolor{level1} 64.0±5.7&
\cellcolor{level5} 95.8±3.8&
\cellcolor{level2} 80.8±4.1&
\cellcolor{level4} 92.6±3.3&
\cellcolor{level4} 94.4±2.1\\

Out-degree&
\cellcolor{level1} 63.0±5.7&
\cellcolor{level5} 96.4±2.5&
\cellcolor{level2} 78.4±5.9&
\cellcolor{level3} 90.6±7.4&
\cellcolor{level4} 93.6±4.5\\

Both degrees&
\cellcolor{level3} \underline{89.4±5.8}&
\cellcolor{level5} 95.0±3.4&
\cellcolor{level2} 80.0±4.3&
\cellcolor{level4} 93.4±2.1&
\cellcolor{level4} 94.4±4.3\\

    \bottomrule
\end{tabular}

\end{small}
\end{center} 

\end{table}

GraphSAGE’s layer-wise update is 
\begin{equation}
    \sigma (\mX\widetilde{\mW_1}+ \mD^{-1}\mA\mX\widetilde{\mW_2}),
\end{equation}
where $\widetilde{\mW_1}$ and $\widetilde{\mW_2}$ are learnable weight matrices.

Comparing with Equation~\ref{eq:out3}, the first two terms correspond exactly to GraphSAGE, except that GCN’s symmetric normalisation replaces GraphSAGE’s row normalisation.
The third term computes the difference between aggregated features from in-neighbors and out-neighbors, i.e., direction imbalance. 
This term contributes little in node classification as MagNet performs poorly on heterophilic graphs \citep{rossiEdgeDirectionalityImproves2023}, thus the best performance occurs when $\sin \alpha \approx 0$.

Experimentally, MagNet’s performance closely mirrors that of GraphSAGE across most datasets, as reported in its original paper \citep{zhangMagNetNeuralNetwork}. The only notable exception is the Telegram dataset introduced in the MagNet study.

To better understand this case, we conducted additional experiments on the Telegram dataset with Dir-GNN~\citep{rossiEdgeDirectionalityImproves2023} and MLP. As shown in Table~\ref{tab:tel_row_norm}, symmetric ("Sym") normalization yields over 90\% accuracy across all feature sets, while row ("Row") normalization reduces accuracy to 38\%--80\%. "None" and directional ("Dir") normalizations perform strongly, exceeding 95\% and 92\%, respectively, underscoring Telegram's high sensitivity to normalization: row normalization impairs performance.

Notably, a simple MLP achieves 89.4\% accuracy using only node degree as a feature, while Dir-GNN \citep{rossiEdgeDirectionalityImproves2023} with row normalisation attains only 38\%. This suggests that the row normalisation used in GraphSAGE may suppress degree information \citep{jiang2025demystifyingmpnnsmessagepassing}, which helps explain MagNet's superior performance on Telegram due to its use of GCN-style normalisation.

\medskip
In sum, without the aforementioned coding error, the performance of MagNet would be similar to that of ChebNet, which is worse than GraphSAGE in most cases.

\subsection{HoloNet: Mixed Weight Coefficients}
\label{sc:holonet}

Despite its sophisticated theoretical motivation, the practical performance of MagNet~\citep{zhangMagNetNeuralNetwork} on directed graphs is underwhelming, casting doubt on whether spectral convolutions can be effectively extended to directed graphs. 
Building on this line of work, the more recent HoloNet model~\citep{kokeholonets} was proposed as a follow-up that aims to address these limitations.
In particular, its ComplexFaberConv operator builds on the Hermitian Laplacian formulation introduced in MagNet \citep{zhangMagNetNeuralNetwork} and reports improved results on several directed graph benchmarks.
Although the HoloNet paper \citep{kokeholonets} reports superior performance across various datasets, only the results on Chameleon and Squirrel rely on the ComplexFaberConv module \citep{holonets2025model}. The remaining datasets are evaluated using FaberNet, which is equivalent to multi-scale learning \citep{jiang2026scaleaware}.

The ComplexFaberConv module in HoloNet \citep{kokeholonets} is based on a flawed implementation (see the issues reported at \url{https://github.com/ChristianKoke/HoloNets/issues}). In this section, we analyze this implementation and explain the underlying reason for the unexpectedly strong performance of the flawed code.

Due to the implementation bug, ComplexFaberConv does not perform the intended multi-order aggregation proposed in HoloNet \citep{kokeholonets}, such as $\mA \mA \mX \mW$ or $\mA^\top \mA^\top \mX \mW$ as the order $k$ increases. Instead, only first-order propagations of the form $\mA \mX \mW$ or $\mA^\top \mX \mW$ are computed. As a result, ComplexFaberConv effectively reduces to a weighted combination of first-order aggregations with different coefficients:
\[
\mA^\top \mX \mW_0 + 2^{-1}\mA^\top \mX \mW_1 + \cdots + 2^{-k}\mA^\top \mX \mW_k.
\]

\begin{figure}[ht]
\vspace{0cm}
  \vskip 0.2in
  \begin{center}
    \centerline{\includegraphics[width=\columnwidth]{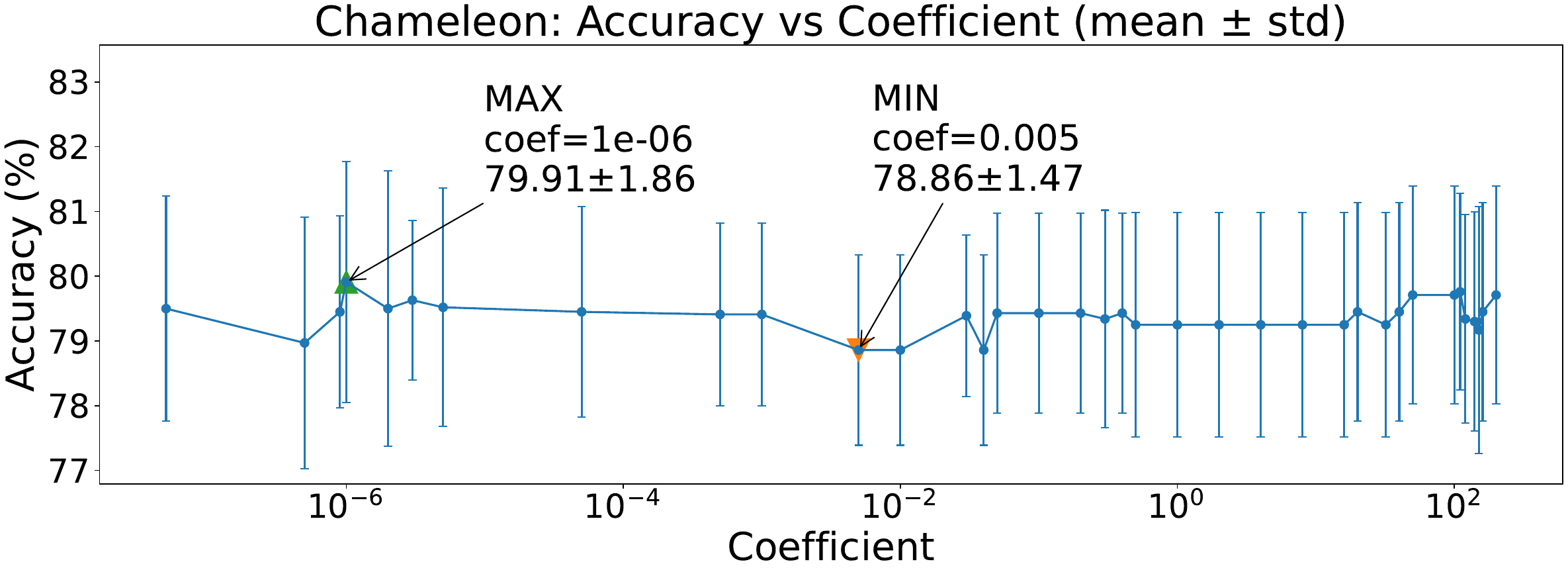}}
    \vspace{1cm} 
    \centerline{\includegraphics[width=\columnwidth]{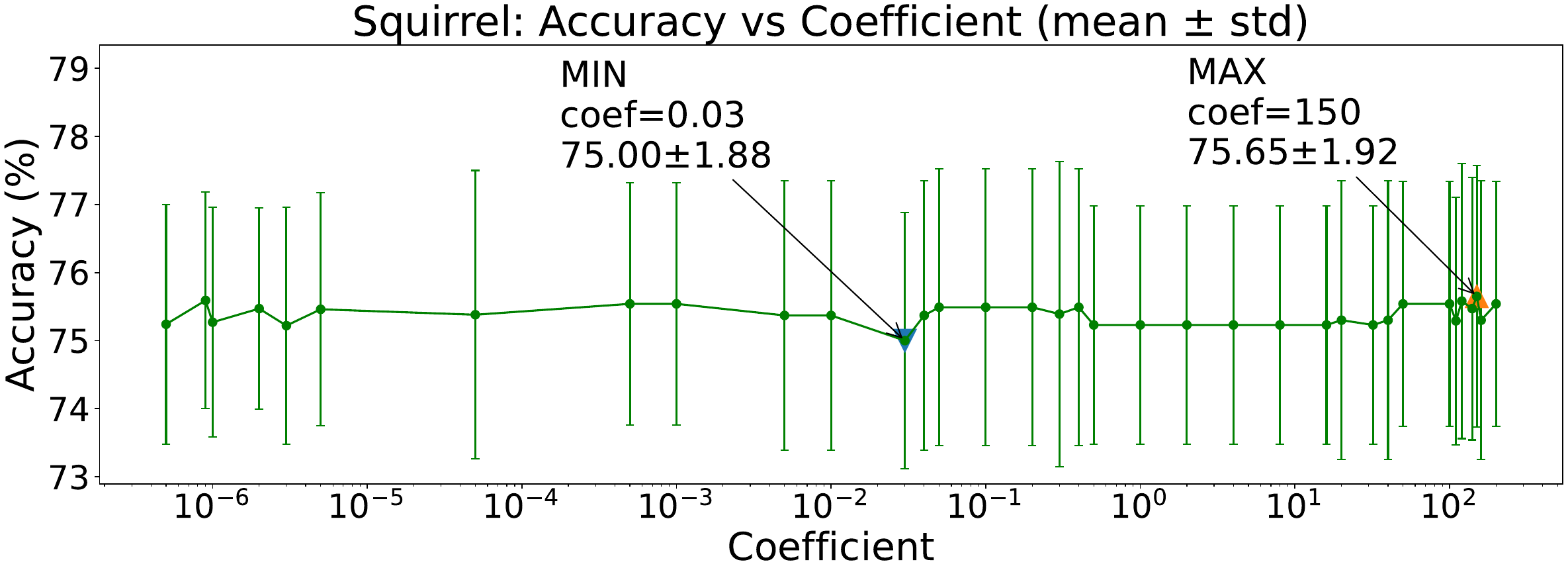}}        
    \caption{
  Accuracy change with different coefficients of Dir-GNN \citep{rossiEdgeDirectionalityImproves2023} on Chameleon and Squirrel datasets. Hyperparameter choices are corresponding to the best parameters in Dir-GNN \citep{rossiEdgeDirectionalityImproves2023}.
    }
    \label{fig_coef_acc}
  \end{center}
\end{figure}

Using $\mL_2$ normalisation can dramatically reduce the influence of coefficients. As shown in Appendix~\ref{sc:l2_norm}, without $\mL_2$ normalisation the accuracy fluctuates drastically—from 19\% to over 75\% on both datasets. With $\mL_2$ normalisation, this sensitivity is substantially reduced: Figure~\ref{fig_coef_acc} shows that accuracy varies only between 78.86\% and 79.91\%. Nevertheless, coefficients can still affect accuracy even under $\mL_2$ normalisation \citep{wangNormFaceL2Hypersphere2017, wangCosFaceLargeMargin2018, dengArcFaceAdditiveAngular2022}. Thus, although the fluctuations are much smaller, coefficients continue to have a non-negligible impact on model performance.

While most models adopt a fixed default coefficient (typically 1), the implementation error in ComplexFaberConv unintentionally mixes multiple coefficients, which can lead to improved empirical performance.


\section{Conclusion}
\label{sc:conclusion}

Recent criticism of Spectral GNNs' Fourier basis claims~\citep{guoPositionSpectralGNNs2025} marks an important step forward, yet the community remains cautiously optimistic due to their empirical effectiveness. Our paper corrects their proof on graph Fourier bases, exposes a second fundamental glitch in polynomial approximations, and reveals the true source of spectral GNN success---equivalence to spatial-domain MPNNs via theoretical reduction or implementation bugs.

For undirected graphs, we show that the widely regarded “powerful” GCN is not a first-order Chebyshev spectral GNN, and its “low-pass” filtering behavior can be fully interpreted through message-passing mechanisms rather than Graph Fourier Transform-based spectral operations. For directed graphs, our analysis of MagNet and HoloNet reveals that empirical evidence no longer supports the claim that spectral methods offer distinct practical advantages.

In summary, Spectral GNNs provide no unique theoretical
or practical benefits.
Spectral GNNs rest on two theoretically flawed steps that, in combination, accidentally recover the MPNN framework — lending them an effectiveness their spectral justification does not warrant.

\section{Alternative Views}

The community's optimism toward Spectral GNNs can be seen from three aspects.

First, as discussed in Sections~\ref{sec:bk_fourier} and \ref{sec:bk_poly}, Spectral GNNs are typically built from two key ingredients: (i) graph Fourier basis and (ii) polynomial approximations. For directed graphs, a common strategy is to construct a Hermitian operator (Section~\ref{sc:bk_spec_direct}) so that a graph Fourier-type basis remains available.

Second, a recent position paper~\citep{guoPositionSpectralGNNs2025} argues that the Fourier bases adopted by many Spectral GNNs lack a clear spectral semantics. Nevertheless, the authors remain optimistic about Spectral GNNs because these models often achieve strong empirical performance.

Third, Spectral GNNs have been extended to directed graphs, and several widely used directed spectral architectures, such as MagNet~\citep{zhangMagNetNeuralNetwork} and HoloNet~\citep{kokeholonets}, have been highly influential, directly and indirectly shaping many follow-up works. More broadly, new variants of Spectral GNNs continue to appear at scale each year, reflecting sustained interest and optimism within the community.

For these reasons, we treat this prevailing optimism---either the belief that Spectral GNNs genuinely ``capture the graph spectrum'' in a meaningful way, or the belief that Spectral GNNs are effective in practice---as our alternative view.
In this position paper, we focus specifically on Spectral GNNs whose constructions are explicitly based on the graph Fourier basis.

\section{Call to Action}
Since spectral GNNs are theoretically flawed, there is no solid reason to expect their ``spectral'' mechanisms to consistently yield superior performance. When strong results appear, they are more plausibly explained by equivalences to simpler MPNNs, or by implementation/training effects rather than any meaningful spectral semantics. Therefore, we call for a more decisive shift in how the community allocates attention and evaluates new spectral claims.
\begin{itemize}
\item \textbf{Stop new research on spectral GNNs for node classifications.}
We argue that Spectral GNNs is theoretically wrong, and more spectral constructions is unnecessary.

\item \textbf{Be suspicious of good performance reported by spectral GNNs.}
When a new spectral GNN reports strong performance, do not take the ``spectral'' explanation at face value. Verify reproducibility and check that the implementation is consistent with the algorithm claimed in the paper.

\item \textbf{Reframe filtering in the hop domain and study hop-invariant operators.}
Low- and high-pass behavior can be characterized more directly in the hop domain through depth-indexed message passing, analogous to shift-invariant filtering in LTI systems. We encourage deeper investigation of hop-invariant formulations and their associated inductive biases, rather than relying on Graph Fourier Transform–based signal processing when its underlying assumptions are not warranted.

\end{itemize}




\bibliography{0_gnn=mpnn}
\bibliographystyle{icml2026}


\newpage
\appendix
\onecolumn
\section{Appendix}
\subsection{Further Discussions}
\label{sec:fur_dis}

In this section, we provide further discussions highlighting fundamental problems with spectral GNNs.

\paragraph{Feature-Driven Tasks}
Classical spectral graph theory \citep{chung1997spectral, hammondWaveletsGraphsSpectral2009a} assumes identical nodes, differing only in their positions within the graph, making graph connectivity the sole focus of analysis. 
Modern node classification tasks, however, involve graphs where node features differ substantially from node to node. Consequently, spectral notions of "low/high frequency" ignore node feature distributions, making Laplacian-based spectral intuition theoretically unsound for learning tasks on graphs with diverse node features.

\paragraph{Eigenvector Instability}
Real-world graphs like citation networks tolerate minor edge/node perturbations without affecting node classification performance. Yet Laplacian eigendecomposition---particularly eigenvector rotations in near-degenerate eigenspaces---exhibits extreme sensitivity to such changes \citep{tao2008eigenvalues, davisRotationEigenvectorsPerturbation1970, hataLocalizationLaplacianEigenvectors2017, noscheseEigenvectorSensitivityGeneral2019}. 
This extreme sensitivity to minor fluctuations contradicts the stability requirements of node classification tasks and undermines spectral GNNs' reliance on eigenvectors as robust graph ``frequencies." Such disconnect suggests spectral methods fundamentally mismatch practical graph learning needs.


\paragraph{Polynomial Choices Are Arbitrary}
Specific polynomial families (Chebyshev in ChebNet \citep{defferrardConvolutionalNeuralNetworks2016}, Bernstein in BernNet \citep{BernNet}) are presented as principled approximations, yet alternatives exist without unique theoretical justification. Notably, higher-order polynomials do not consistently improve performance \citep{BernNet}. Low-order approximations like GCN \citep{kipfSemiSupervisedClassificationGraph2017}---a first-order Chebyshev simplification---often outperform more complex spectral models like ChebNet \citep{defferrardConvolutionalNeuralNetworks2016, zhangMagNetNeuralNetwork}. Even ChebNet typically uses small $K=2$, as larger $K$ degrades performance. 
These observations suggest that the practical success of many ``spectral'' GNNs may be better explained by their induced spatial aggregation behavior (and the associated regularization/inductive bias) than by a literal Fourier interpretation.

\medskip
Overall, while Laplacian eigenvectors mathematically decompose operators derived from graph connectivity, interpreting them as a classical Fourier basis with frequency semantics for the graph itself lacks justification. 
Spectral terminology thus provides no principled guidance for architecture design or interpretability claims about ``frequencies'' on graphs.

\subsection{Low Pass Filter in Discrete Domain}
\label{sec:dft_low_pass}

Consider the Linear Time-Invariant (LTI) Systems with input  $x[n]$ and output
\[
y[n] = x[n] +x[n-1].
\]
Its impulse response is
\[
h[n] = \delta[n] + \delta[n-1], 
\]
where $\delta[n]$ is Kronecker delta function.

The discrete‑time Fourier transform (DTFT) of $h[n]$ gives the frequency response
\begin{equation}
    \label{eq:h(w)}
    H(\omega)=\sum_{n=-\infty}^{\infty} h[n]\,e^{-i\omega n}=1+e^{-i\omega}.
\end{equation}
For an input $x[n]=e^{i\omega n}$, the output is
\[
y[n]=H(\omega)\,e^{i\omega n},
\]
so each basis function $e^{i\omega n}$ is scaled by $H(\omega)$.

\textbf{Magnitude Response}

Rewrite Equation~\ref{eq:h(w)}, we get:
\[
H(\omega)=e^{-i\omega/2}\left(e^{i\omega/2}+e^{-i\omega/2}\right)
=2\cos\!\left(\frac{\omega}{2}\right)e^{-i\omega/2}.
\]
Thus,
\[
|H(\omega)|=2\left|\cos\!\left(\frac{\omega}{2}\right)\right|.
\]

\textbf{Low-Pass Property} 

At $\omega=0$,
$|H(0)|=2 \qquad \text{(maximum gain)}.$

At $\omega=\pi$,
$|H(\pi)|=0 \qquad \text{(zero gain)}.$

Moreover, $|H(\omega)|$ decreases monotonically from $2$ to $0$ as $\omega$ increases from $0$ to $\pi$.

\subsection{Influence of $\mL_2$ Normalization on Stability}
\label{sc:l2_norm}

\begin{figure}[ht]
\vspace{0cm}
  \vskip 0.2in
  \begin{center}
    \centerline{\includegraphics[width=\columnwidth]{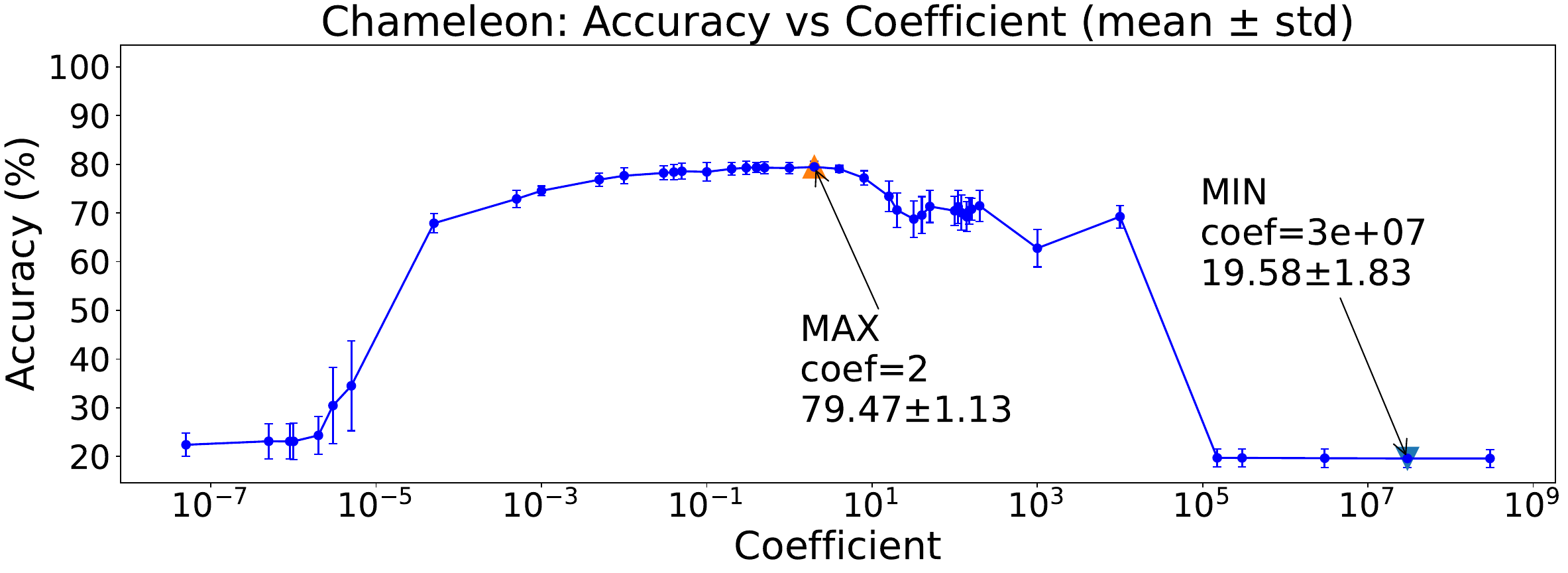}}
    \vspace{1cm} 
    \centerline{\includegraphics[width=\columnwidth]{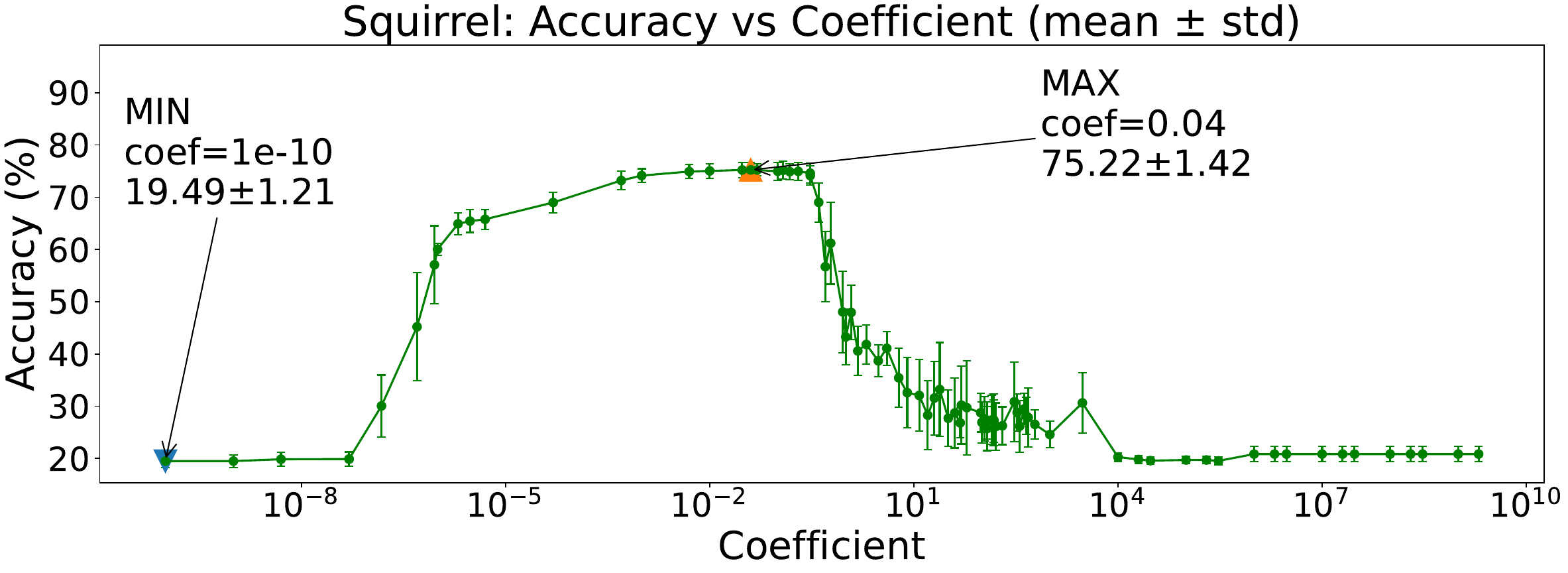}}        
    \caption{
  Accuracy change with different coefficients of Dir-GNN \citep{rossiEdgeDirectionalityImproves2023} on Chameleon and Squirrel datasets. Hyperparameter choices are corresponding to the best parameters in Dir-GNN, except that without $L_2$ normalisation.
    }
    \label{fig_coef_acc_n0}
  \end{center}
\end{figure}

When $\mL_2$ normalization is applied to node features by rescaling $\mX$ such that each node vector has unit $\mL_2$ norm. normalizes each node's feature vector, which stabilizes learning and reduces sensitivity to the scale (coefficients) of the subsequent linear weights. 
As shown in Figure~\ref{fig_coef_acc_n0}, without this stabilization the accuracy varies dramatically: on \textsc{Chameleon} it fluctuates from 19.58\% to 79.47\%, and on \textsc{Squirrel} from 19.49\% to 79.22\%.
\citet{wangNormFaceL2Hypersphere2017} discuss the necessity of $\mL_2$ normalization to improve stability.

\end{document}